\documentclass[letterpaper, 10 pt, conference]{ieeeconf}

\IEEEoverridecommandlockouts                              

\usepackage{multirow}
\usepackage{epstopdf}
\usepackage{dblfloatfix}
\usepackage{cite}
\usepackage{wrapfig}
\usepackage{listings}
\usepackage{float}
\usepackage[dvipdfmx]{graphicx}
\usepackage{amssymb}
\usepackage{latexsym}
\usepackage{amsfonts}
\usepackage{url}
\usepackage{comment}
\usepackage[linesnumbered,ruled,vlined]{algorithm2e}
\usepackage{algpseudocode}
\usepackage{amsmath}
\usepackage{booktabs}
\usepackage{dcolumn}

{}

\newcommand{\figTeaser}{
\begin{figure}[t]
    \centering
    \includegraphics[width=\linewidth]{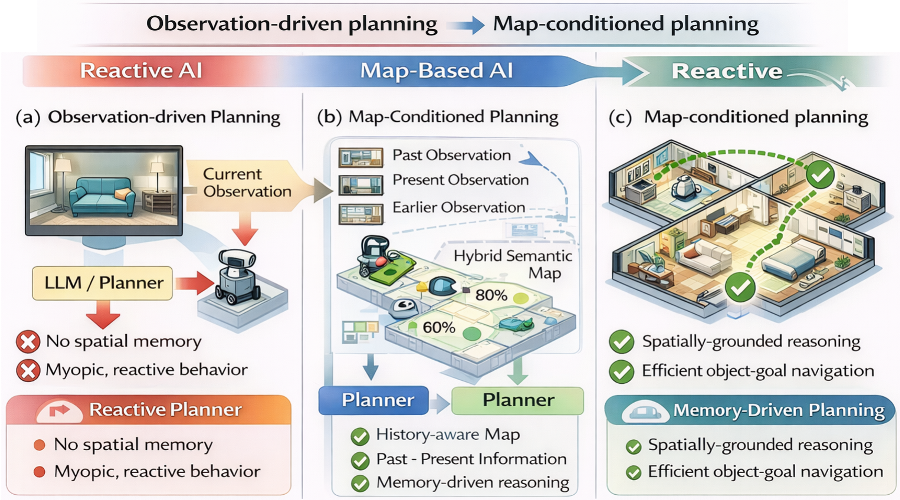}
    \caption{Concept of the proposed memory-driven planning. Compared to (a) traditional observation-driven planning, which often results in myopic and redundant behaviors due to the lack of spatial memory, our approach transitions toward (b) map-conditioned planning. By integrating a hybrid semantic map with LLM-based reasoning, the agent achieves (c) efficient object-goal navigation through spatially-grounded decision-making and systematic exploration.}
    \label{fig:concept}
\end{figure}
}

\newcommand{\figSystemArchitecture}{
\begin{figure}[t]
    \centering
    \includegraphics[width=\linewidth]{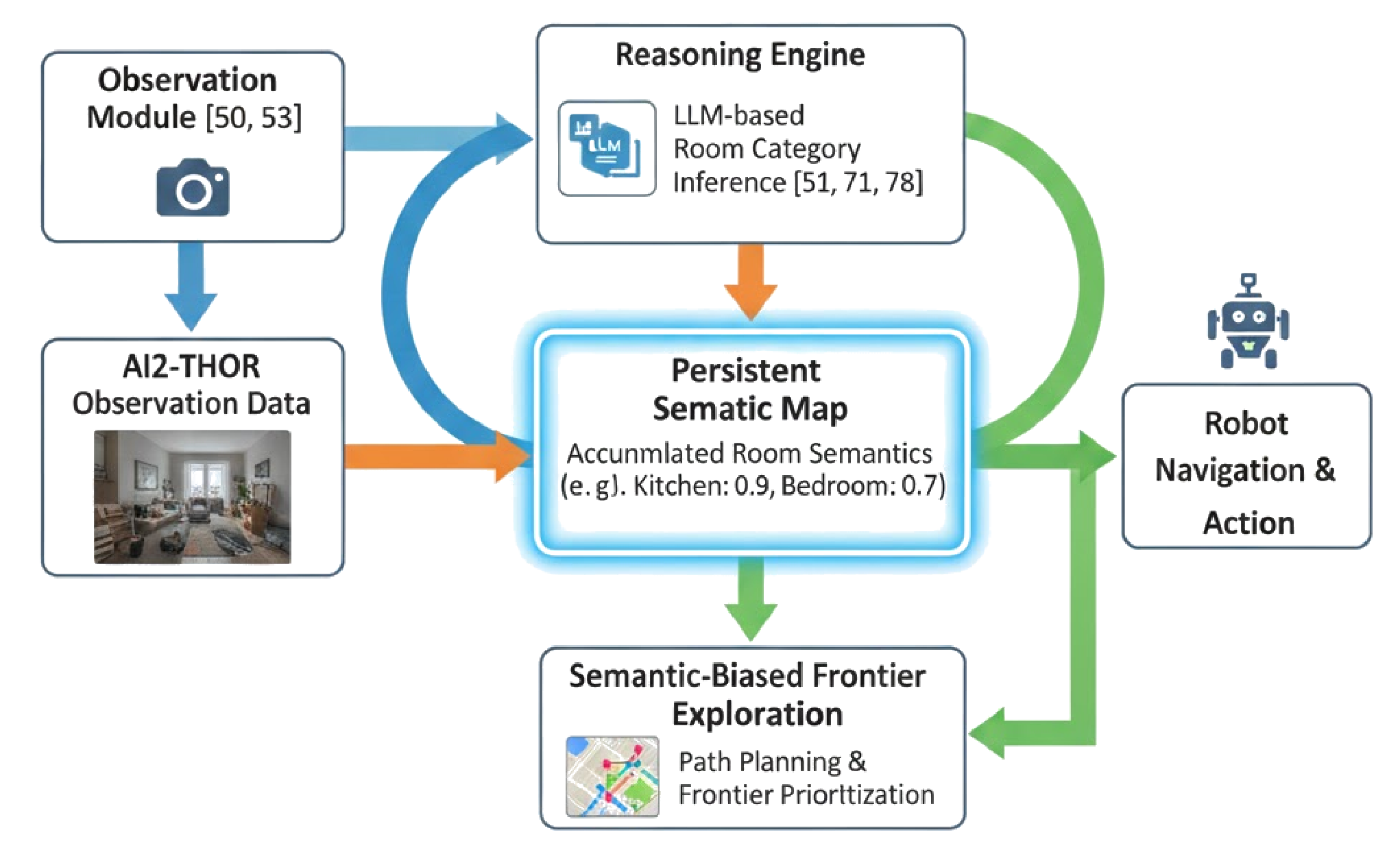}
    \caption{Architecture of the proactive exploration system. It integrates semantic mapping, spatial correlation reasoning, and a strategic path planner.}
    \label{fig:system_architecture}
\end{figure}
}

\newcommand{\figEnvironmentMaps}{
\begin{figure*}[t]
    \centering
    \begin{minipage}[b]{0.19\linewidth}
        \centering
        \includegraphics[width=\linewidth]{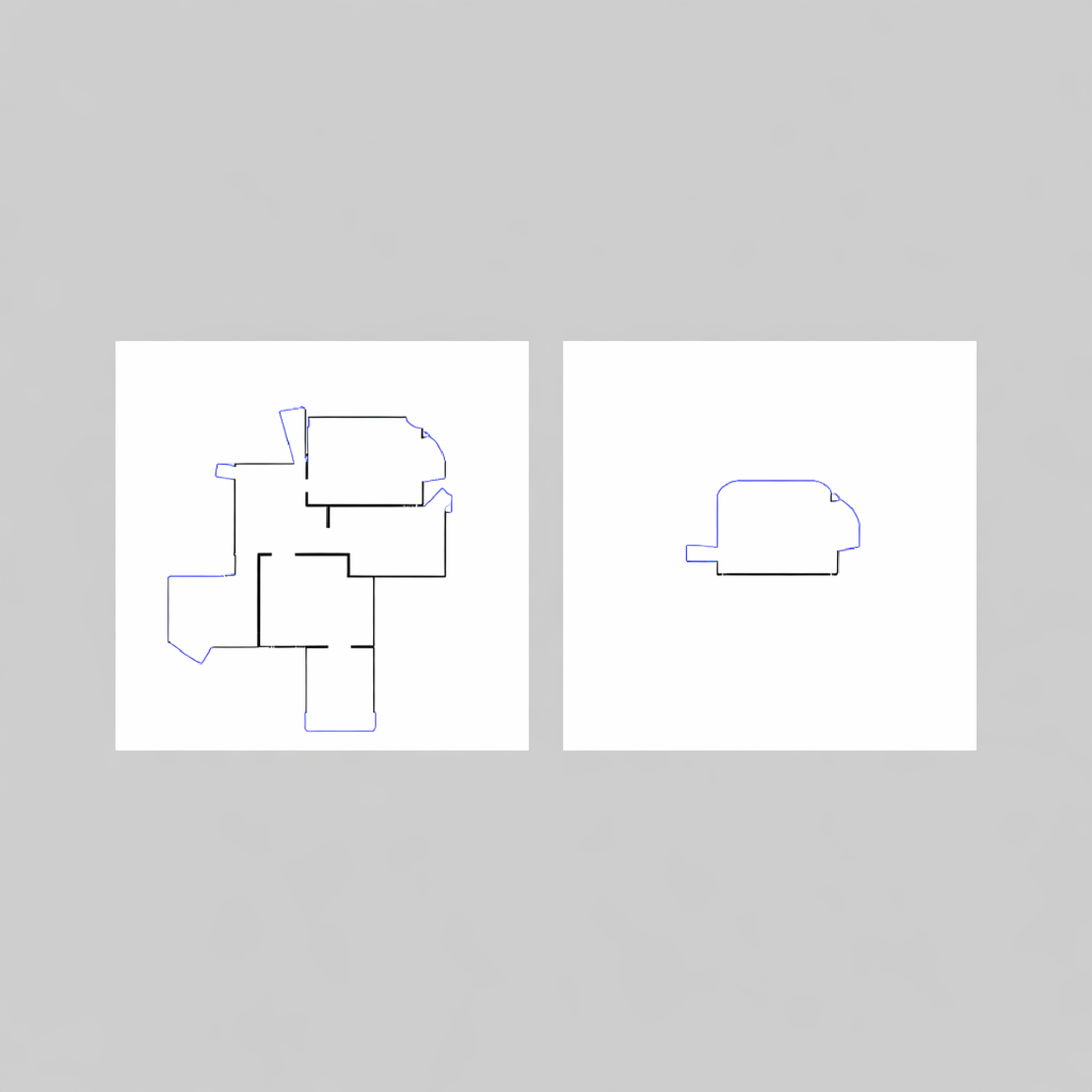} \\ (a) Floor plan
    \end{minipage}
    \begin{minipage}[b]{0.19\linewidth}
        \centering
        \includegraphics[width=\linewidth]{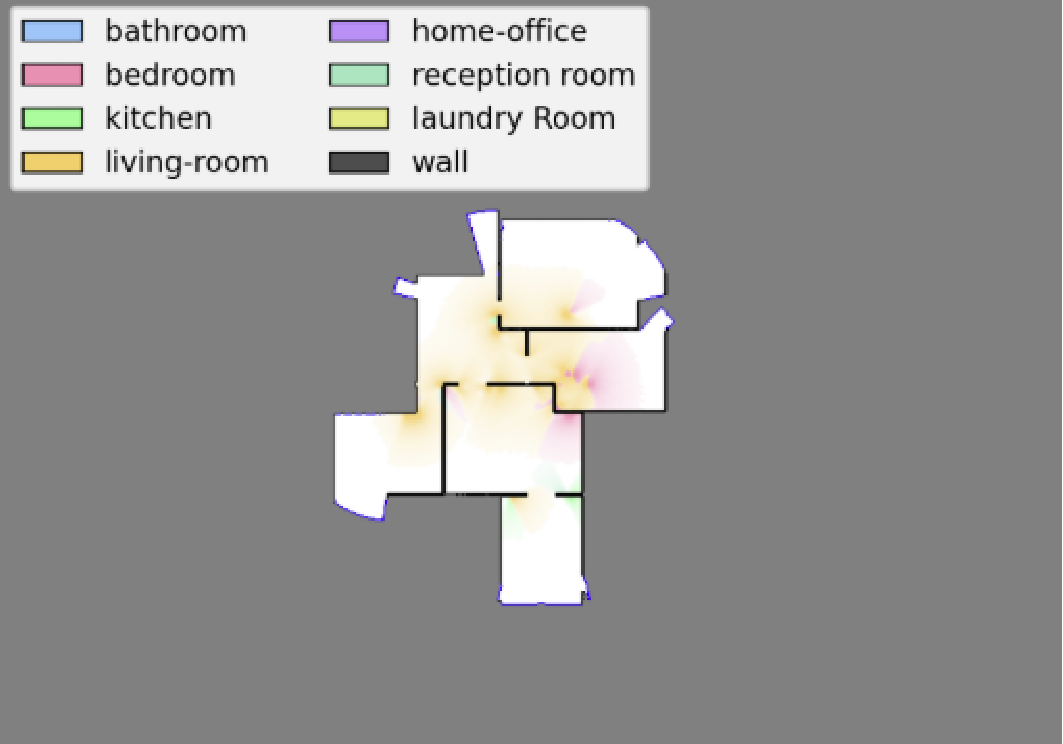} \\ (b) Semantic map
    \end{minipage}
    \begin{minipage}[b]{0.19\linewidth}
        \centering
        \includegraphics[width=\linewidth]{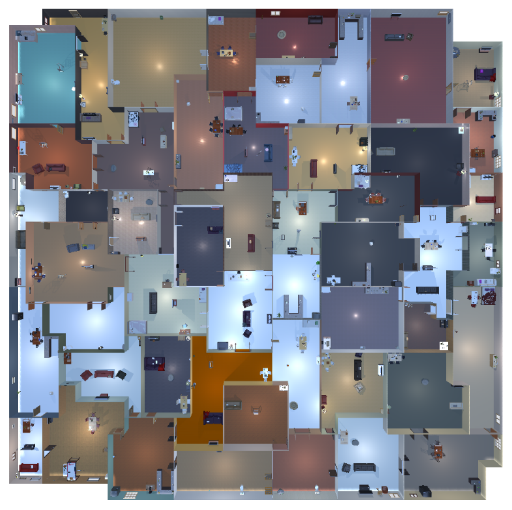} \\ (c) Layout A
    \end{minipage}
    \begin{minipage}[b]{0.19\linewidth}
        \centering
        \includegraphics[width=\linewidth]{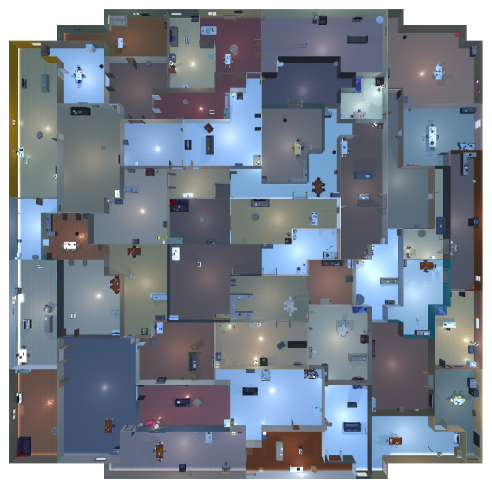} \\ (d) Layout B
    \end{minipage}
    \begin{minipage}[b]{0.19\linewidth}
        \centering
        \includegraphics[width=\linewidth]{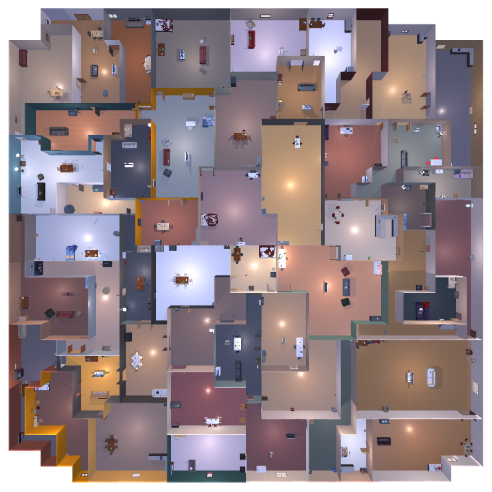} \\ (e) Layout C
    \end{minipage}
    \caption{Overview of simulation environments. (a-b) show the spatial and semantic definitions, while (c-e) present the top-down views of various test scenes.}
    \label{fig:env_maps}
\end{figure*}
}

\newcommand{\figAgentObservationsLiving}{
\begin{figure*}[t]
    \centering
    \includegraphics[width=0.24\linewidth]{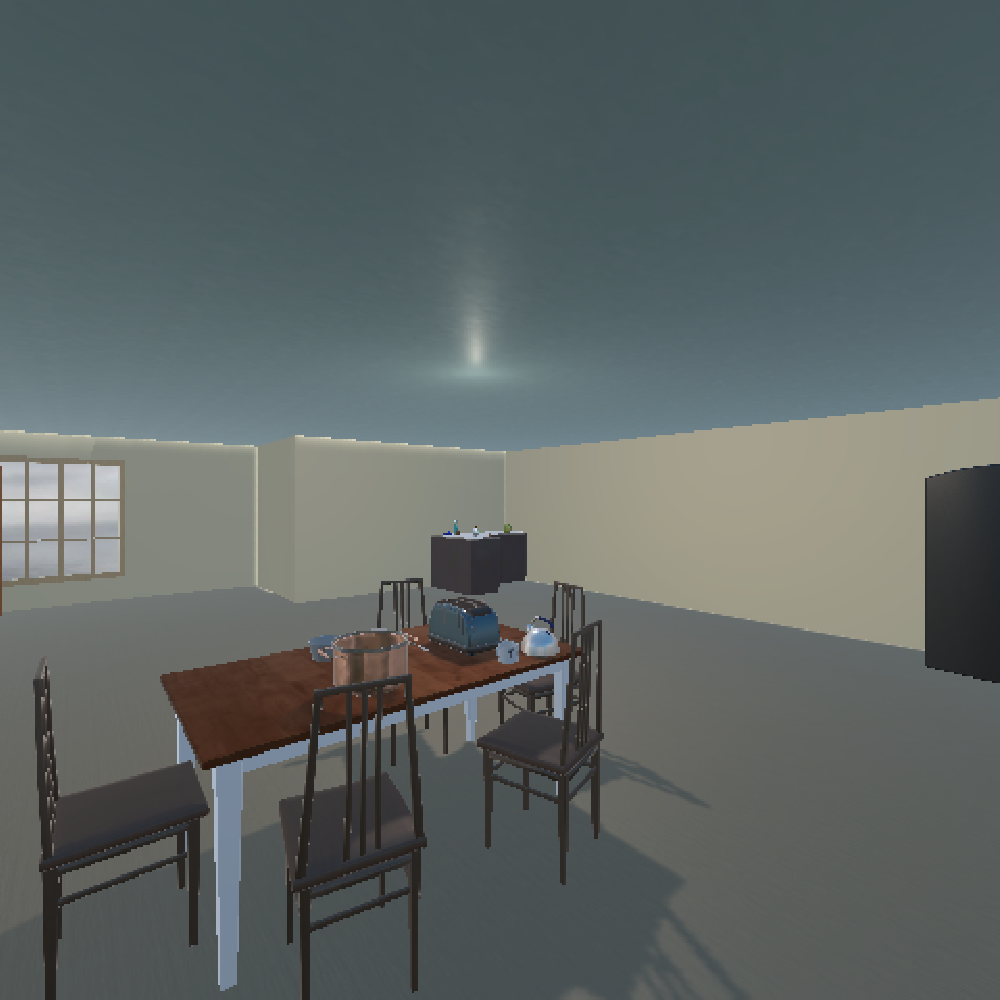}
    \includegraphics[width=0.24\linewidth]{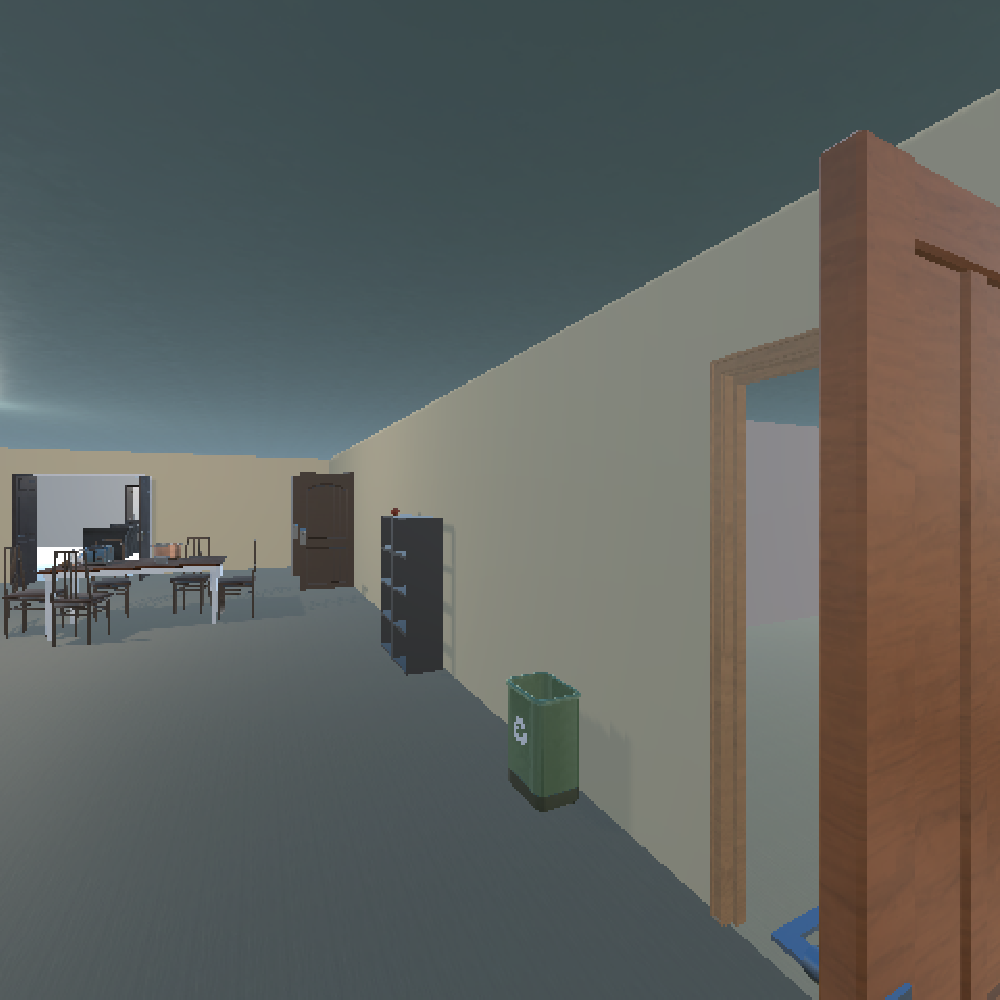}
    \includegraphics[width=0.24\linewidth]{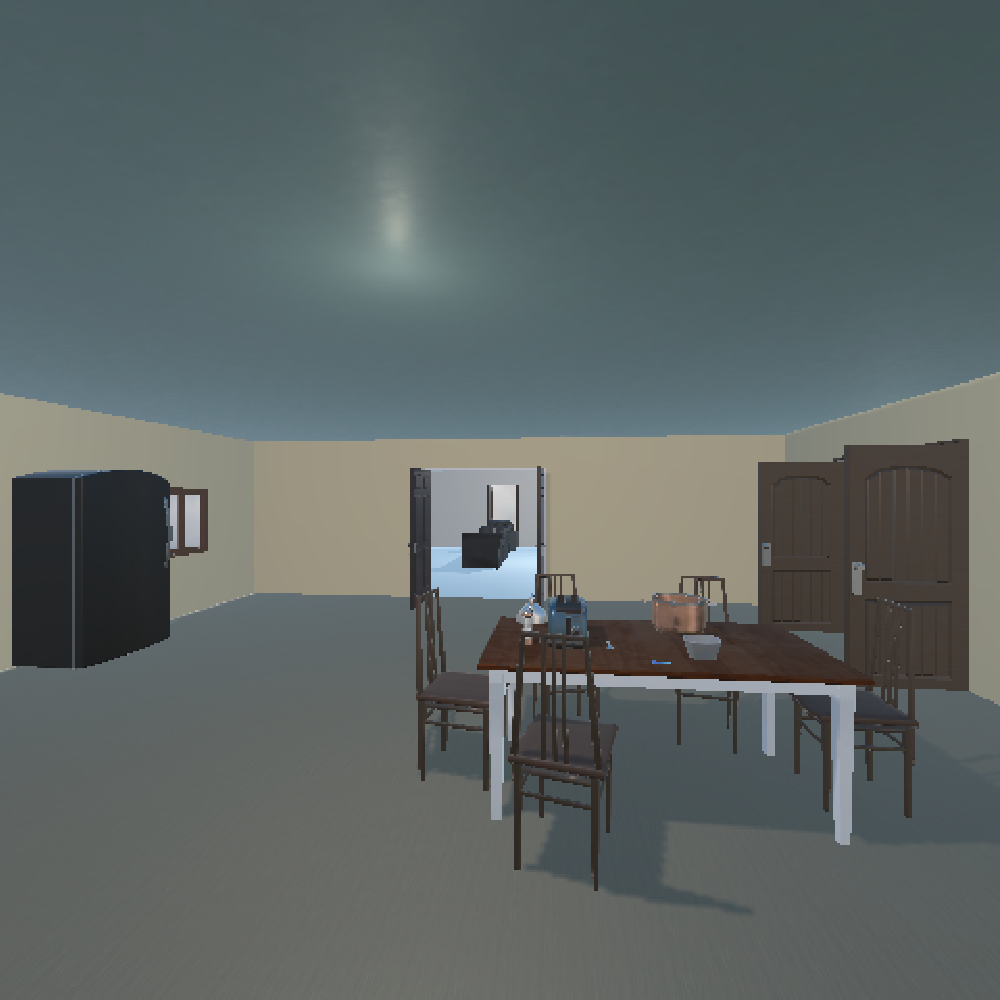}
    \includegraphics[width=0.24\linewidth]{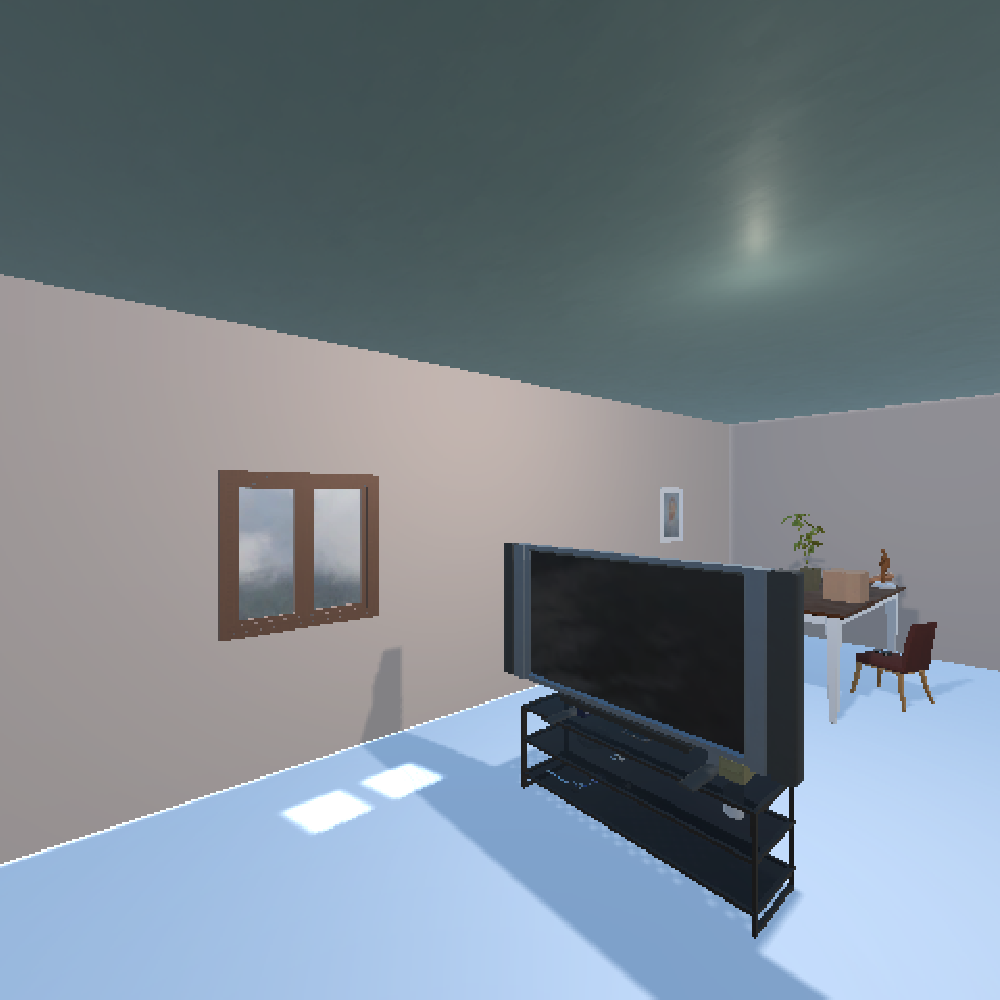} \\[1mm]
    \includegraphics[width=0.24\linewidth]{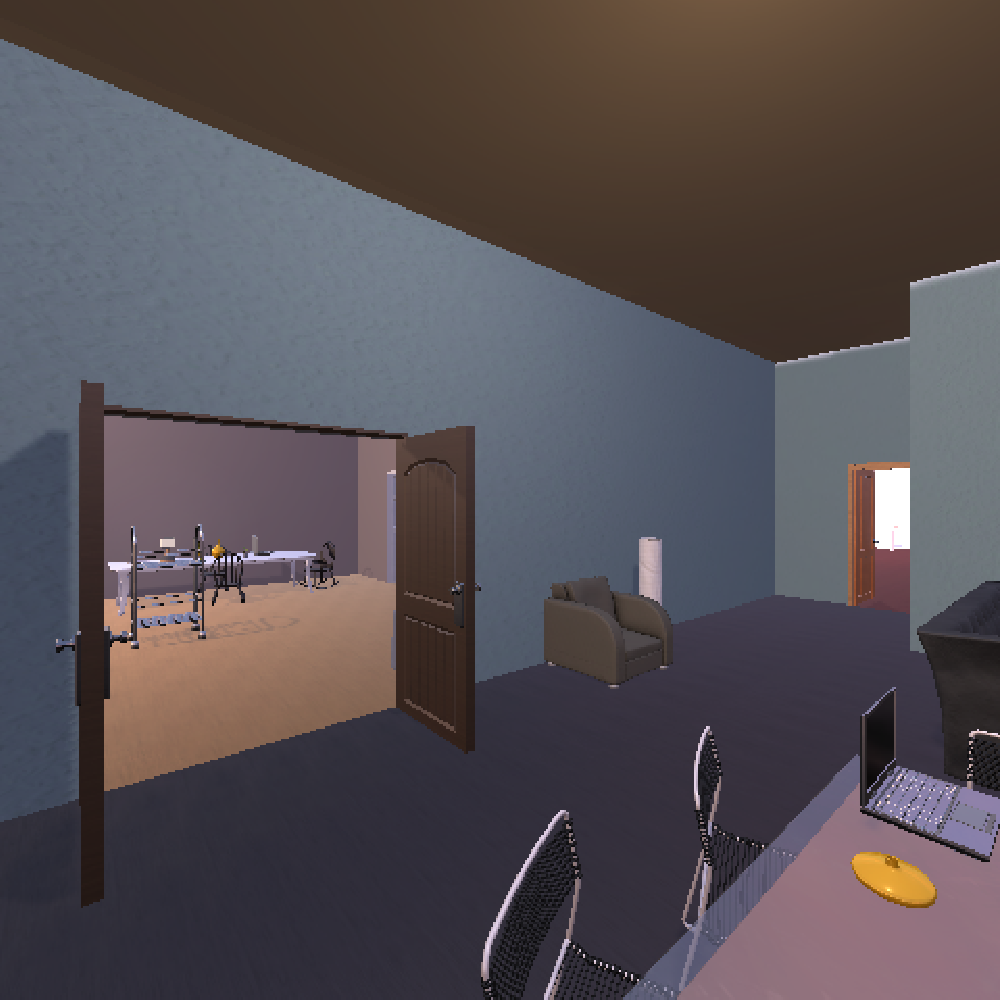}
    \includegraphics[width=0.24\linewidth]{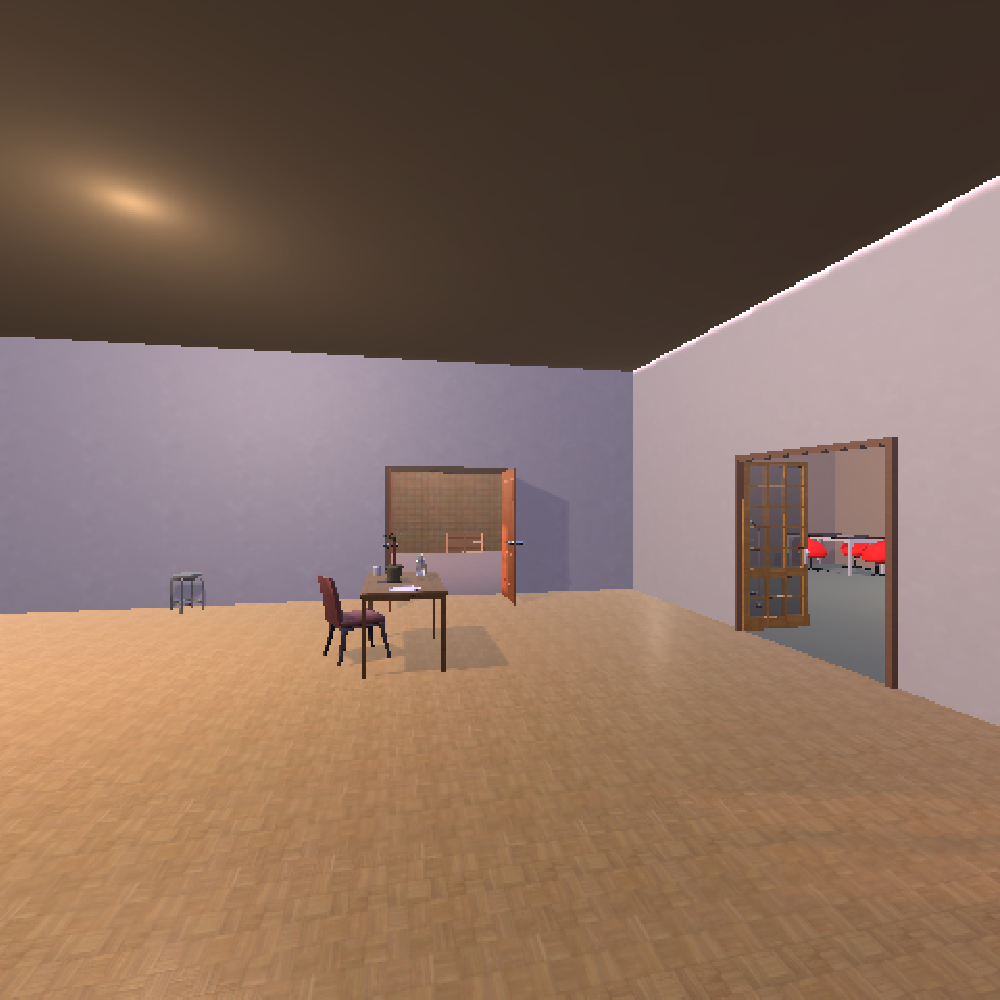}
    \includegraphics[width=0.24\linewidth]{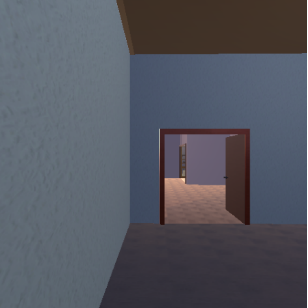}
    \includegraphics[width=0.24\linewidth]{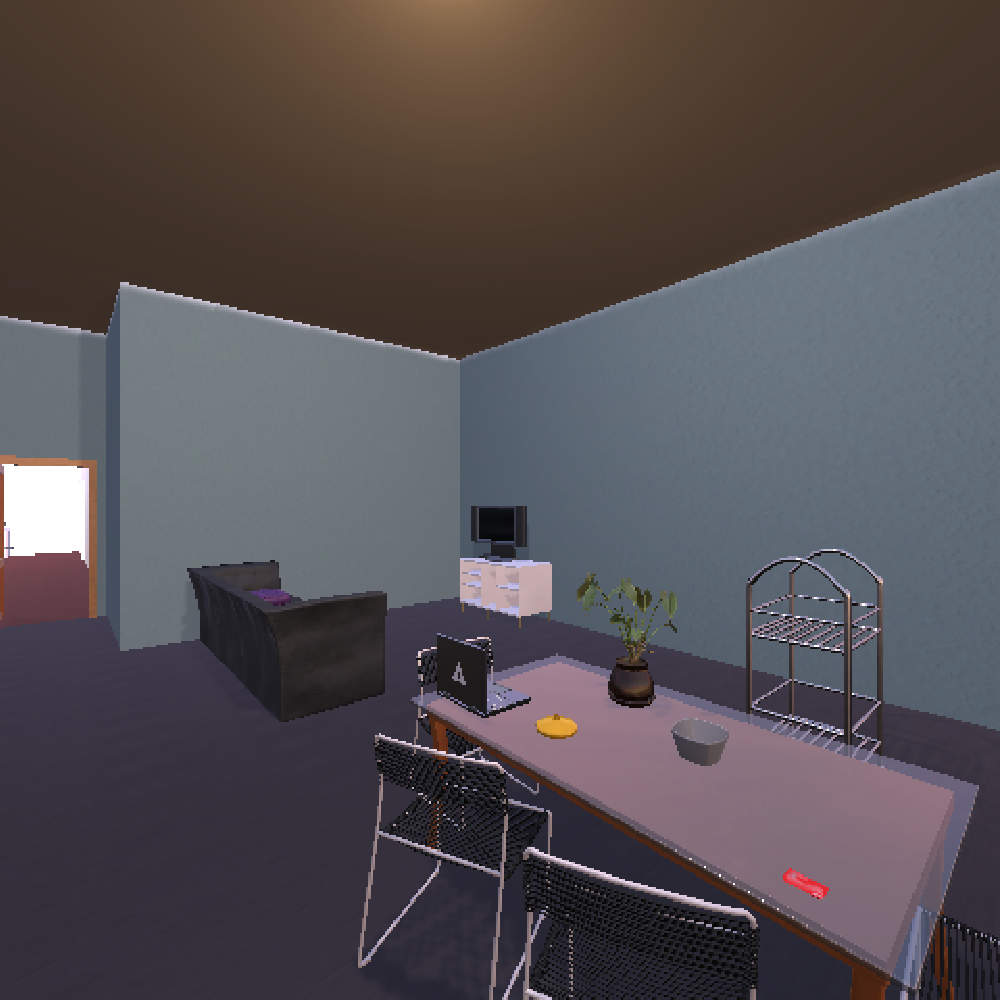}
    \caption{First-person observations in living and sleeping quarters. These images represent the visual input for the object detection and localization modules.}
    \label{fig:obs_living}
\end{figure*}
}

\newcommand{\figExperimentalResults}{
\begin{figure*}[t]
\scriptsize
    \centering
    \begin{minipage}[b]{0.24\linewidth}
        \centering
        \includegraphics[width=\linewidth]{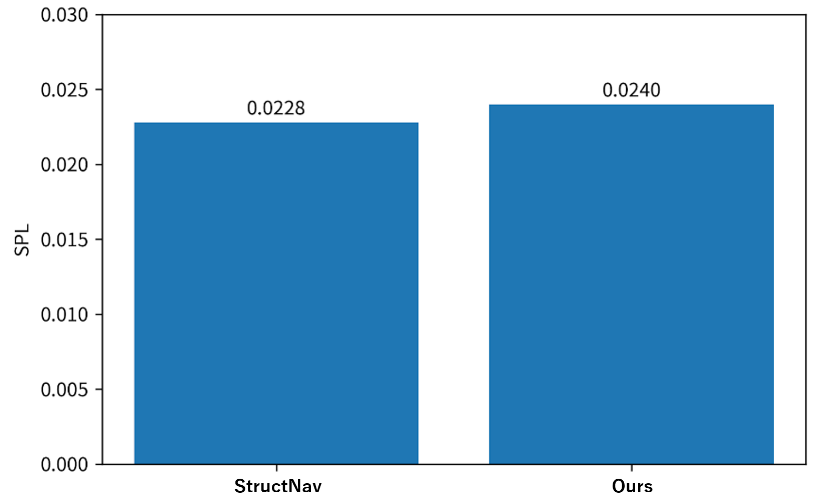} \\ (a) Overall success
    \end{minipage}
    \begin{minipage}[b]{0.24\linewidth}
        \centering
        \includegraphics[width=\linewidth]{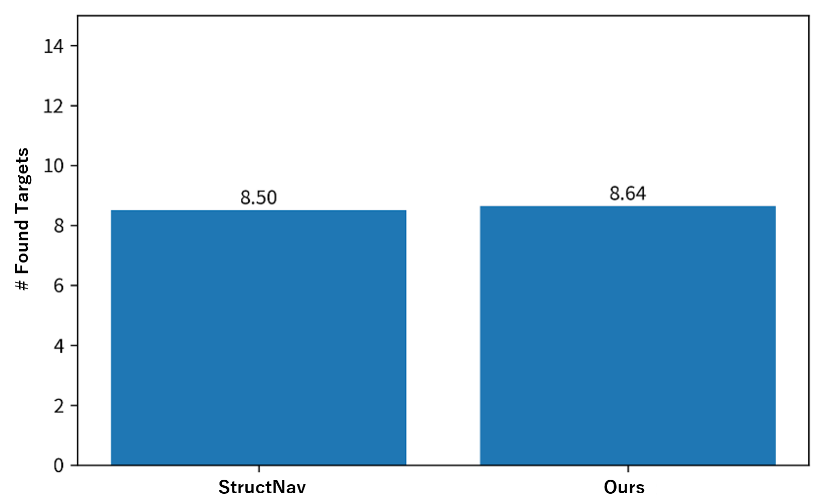} \\ (b) Metric efficiency
    \end{minipage}
    \begin{minipage}[b]{0.24\linewidth}
        \centering
        \includegraphics[width=\linewidth]{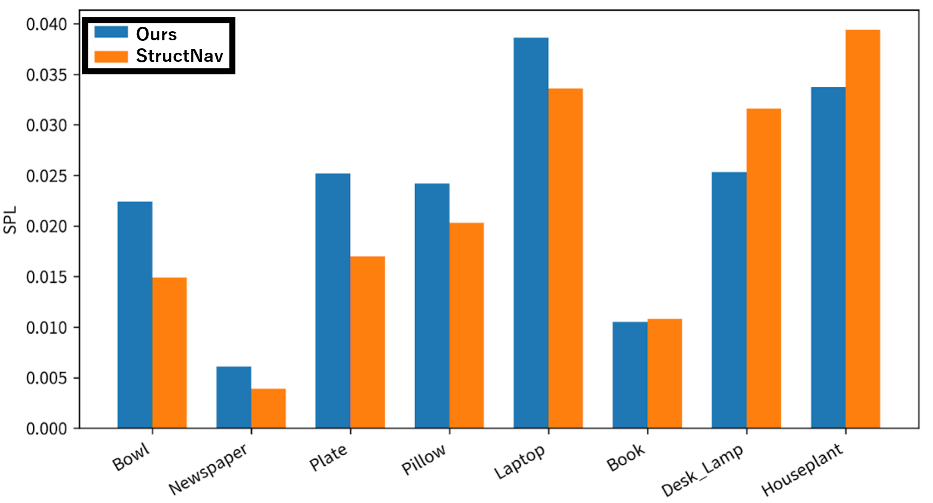} \\ (c) Step comparison
    \end{minipage}
    \begin{minipage}[b]{0.24\linewidth}
        \centering
        \includegraphics[width=\linewidth]{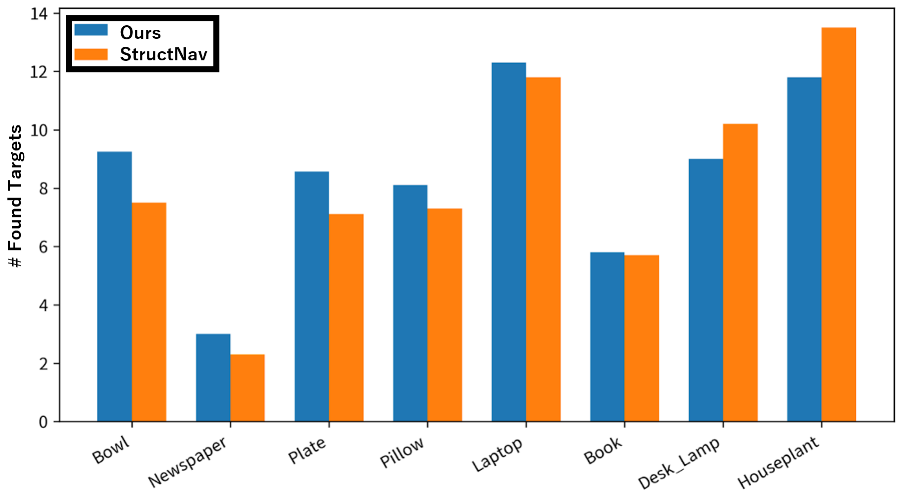} \\ (d) Detailed analysis
    \end{minipage}
    \caption{Quantitative performance evaluation. Our proposed method significantly outperforms baseline approaches in both search speed and path efficiency.}
    \label{fig:results}
\end{figure*}
}

\begin{document}

\title{\LARGE \bf
From Reactive to Map-Based AI: Tuned Local LLMs for Semantic Zone Inference in Object-Goal Navigation
}

\author{Yudai Noda$^{1}$ and Kanji Tanaka$^{1}$
\thanks{*This work was supported in part by JSPS KAKENHI.}
\thanks{$^{1}$Yudai Noda and Kanji Tanaka are with the Graduate School of Engineering, University of Fukui, Japan.
        {\tt\small tnkknj@u-fukui.ac.jp}}%
}

\maketitle
\thispagestyle{empty}
\pagestyle{empty}

\begin{abstract}
Object-Goal Navigation (ObjectNav) requires an agent to find and navigate to a target object category in unknown environments. While recent Large Language Model (LLM)-based agents exhibit zero-shot reasoning, they often rely on a "reactive" paradigm that lacks explicit spatial memory, leading to redundant exploration and myopic behaviors. To address these limitations, we propose a transition from reactive AI to "Map-Based AI" by integrating LLM-based semantic inference with a hybrid topological-grid mapping system. Our framework employs a fine-tuned Llama-2 model via Low-Rank Adaptation (LoRA) to infer semantic zone categories and target existence probabilities from verbalized object observations. In this study, a "zone" is defined as a functional area described by the set of observed objects, providing crucial semantic co-occurrence cues for finding the target. This semantic information is integrated into a topological graph, enabling the agent to prioritize high-probability areas and perform systematic exploration via Traveling Salesman Problem (TSP) optimization. Evaluations in the AI2-THOR simulator demonstrate that our approach significantly outperforms traditional frontier exploration and reactive LLM baselines, achieving a superior Success Rate (SR) and Success weighted by Path Length (SPL).
\end{abstract}

\section{INTRODUCTION}
\label{sec:introduction}

\subsection{Motivation}
In recent years, Object-Goal Navigation (ObjectNav) has emerged as a fundamental challenge in service robotics, requiring an agent to locate and navigate to a target object category within an unknown indoor environment~\cite{chaplot2020object}. Efficient exploration in such settings is inherently difficult due to the lack of prior spatial knowledge and the infinite search space. Traditionally, these tasks have been addressed using geometric exploration strategies, such as frontier exploration~\cite{yamauchi1997frontier}, or deep reinforcement learning (DRL) methods~\cite{chaplot2020ans}.

\figTeaser

However, these approaches often lack semantic ``commonsense''---the ability to predict, for instance, that a ``kettle'' is more likely to be found in a ``zone'' containing a ``stove''~\cite{anderson2018evaluation}. In our framework, such a location is described not by architectural room labels, but by the aggregate set of observed objects, which serves as a powerful cue for identifying target-related contexts.

The integration of Large Language Models (LLMs) has introduced powerful zero-shot reasoning capabilities to robotics~\cite{yao2022react}. Yet, as illustrated in Fig.~\ref{fig:concept}, most existing LLM-based navigation agents rely on a "reactive" paradigm, where actions are generated sequentially based only on current observations~\cite{dorbala2024lgx}. This often leads to sub-optimal, myopic behaviors, such as repetitive visits to the same area and a lack of systematic coverage. To address these limitations, this study proposes a transition from reactive AI to ``Map-Based AI,'' which externalizes exploration history and spatial structure into a hybrid map to ensure both semantic efficiency and exploration completeness~\cite{chen2023how}.

\subsection{Problem Statement}
Classical geometric-based exploration focuses on maximizing map coverage by targeting frontiers (boundaries between known and unknown space)~\cite{elfes1989occupancy}. While effective for mapping, these methods are ``semantic-blind'' and may exhaustively search irrelevant zones, leading to excessive path lengths~\cite{savva2019habitat}. Conversely, while LLMs provide semantic guidance, without an explicit spatial memory (map), they struggle with long-term consistency~\cite{brown2020gpt3}. The fundamental problem lies in the lack of a framework that seamlessly integrates high-level semantic reasoning with low-level metric and topological representations of the environment, where locations are defined and distinguished by their functional object clusters~\cite{chaplot2020topological}.

\subsection{Contribution}
To bridge the gap between reactive reasoning and systematic mapping, this paper makes the following contributions:
\begin{enumerate}
\item \textbf{LLM-based Semantic Zone Inference:} We propose a method to infer semantic zone categories and target existence probabilities by verbalizing observed object sets using a fine-tuned Llama-2 model via Low-Rank Adaptation (LoRA)~\cite{hu2021lora}. We introduce the concept of a ``zone'' as a spatial unit described by its constituent objects, which provides a more robust cue for ObjectNav than traditional room labels.
\item \textbf{Hybrid Topological-Grid Mapping:} We implement a dual-layer mapping system that manages space as a graph of nodes (zones) and edges (transitions), allowing the robot to perform high-level planning over semantic contexts rather than just geometric coordinates~\cite{thrun2005probabilistic}.
\item \textbf{Empirical Validation in AI2-THOR:} We demonstrate through extensive simulations that our map-based approach significantly outperforms traditional frontier exploration and reactive LLM baselines in terms of Success weighted by Path Length (SPL)~\cite{deitke2020robothor}.
\end{enumerate}

\section{RELATED WORK}
\label{sec:related_work}

\subsection{Visual Semantic Navigation}
Visual Semantic Navigation, particularly Object-Goal Navigation (ObjectNav), has been extensively studied as a task requiring both visual perception and semantic reasoning~\cite{chaplot2020object,anderson2018evaluation}. Early methods primarily relied on end-to-end Deep Reinforcement Learning (DRL) to learn direct mappings from pixels to actions~\cite{savva2019habitat,sutton2018reinforcement}. However, these methods often struggle with sample efficiency and generalization to unseen environments.

To mitigate these issues, modular approaches have been proposed, which decompose the task into mapping, goal prediction, and path planning~\cite{chen2023how}. While these modular systems improve reliability, their semantic priors are often restricted by the diversity of the training datasets~\cite{chang2017matterport3d,xia2018gibson,kolve2017ai2thor}.

\subsection{LLMs for Robotic Decision Making}
The emergence of Large Language Models (LLMs) has provided a new paradigm for zero-shot robotic planning~\cite{brown2020gpt3,wei2022cot}. Recent works have demonstrated that LLMs can act as high-level planners by decomposing natural language instructions into a sequence of executable sub-goals~\cite{ahn2022saycan}. In the context of ObjectNav, several studies have employed LLMs to generate the next action based on verbalized visual observations~\cite{dorbala2024lgx}.

However, most of these methods are ``reactive,'' meaning they lack an explicit spatial memory~\cite{yao2022react}. As highlighted in our previous analysis, this reactive nature often leads to ``myopic'' behaviors, such as local loops or redundant exploration of already visited zones, due to the absence of global context~\cite{sutton1991dyna}.

\subsection{Graph-based Mapping and Topological Navigation}
To maintain global consistency, graph-based topological maps have long been used in robotics to represent environments as sets of nodes and edges~\cite{thrun2005probabilistic}. Topological maps offer a more compact representation compared to dense metric grids~\cite{elfes1989occupancy}, facilitating long-term planning and loop closure~\cite{murartal2015orbslam}.

Recently, efforts have been made to augment these graphs with semantic information (e.g., semantic labels)~\cite{mccormac2017semanticfusion,stuckler2014semantic}. Our work extends this direction by using an LLM to dynamically infer the semantic properties of topological nodes (zones) based on discovered object sets~\cite{yao2022react}. In our framework, a zone is defined not by physical boundaries, but by its constituent object clusters, which serve as a semantic signature for the location. This allows the robot to perform ``semantic-topological reasoning,'' bridging the gap between low-level sensorimotor control and high-level commonsense logic~\cite{chaplot2020topological,sunderhauf2017meaningful}.



\figSystemArchitecture

\section{PROPOSED METHOD}
\label{sec:architecture}

\subsection{System Overview}
The proposed framework is designed to transition from a reactive ``observation-to-action'' paradigm to a structured ``map-based reasoning'' paradigm~\cite{yao2022react}. The system architecture, illustrated in Fig.~\ref{fig:system_architecture}, follows a decoupled design consisting of two primary modules: the \textit{Decision-Making Module} (DMM) and the \textit{Environment Interaction Module} (EIM)~\cite{kaelbling1998planning}. 

The EIM handles low-level tasks, including controlling the agent within the AI2-THOR simulator~\cite{kolve2017ai2thor}, performing 360-degree panoramic scans, and pre-processing visual data into semantic labels~\cite{zhou2022detic}. In contrast, the DMM manages high-level cognitive tasks, such as maintaining the hybrid topological-grid map~\cite{thrun2005probabilistic}, performing LLM-based semantic zone inference~\cite{hu2021lora}, and executing global path planning via the A* algorithm~\cite{hart1968astar} and TSP-based optimization~\cite{weihs2021pddl}. This asynchronous communication via a file-based Inter-Process Communication (IPC) ensures that the high-latency LLM inference does not bottleneck the real-time perception and control loop~\cite{sutton1991dyna}.

\subsection{Perception Layer}
The perception layer transforms raw RGB-D observations into structured semantic information required for mapping and reasoning. This involves semantic correlation analysis and spatial noise filtering.

\subsubsection{Semantic Scoring via SBERT}
To evaluate the relevance of observed objects to the search target, we utilize Sentence-BERT (SBERT) to compute semantic embeddings. Let $w_{target}$ be the embedding of the target object category and $w_{obs}$ be the embedding of a detected object label. The semantic similarity score $S$ is defined as the cosine similarity:
\begin{equation}
S(w_{target}, w_{obs}) = \frac{w_{target} \cdot w_{obs}}{|w_{target}| |w_{obs}|}
\end{equation}
This allows the agent to prioritize frontiers near objects that are functionally related to the target (e.g., a ``stove'' when searching for a ``kettle'') based on the linguistic commonsense embedded in SBERT.

\subsubsection{Spatial and Visual Filtering}
To ensure the reliability of the semantic map, we implement a multi-stage filter. A detected object instance $o_i$ is only integrated into the map if it satisfies both the visual and spatial constraints:
\begin{itemize}
\item \textbf{Pixel Constraint:} $Area(o_i) \ge \tau_{pixel}$ (e.g., 400 pixels), ensuring the object is large enough for reliable identification.
\item \textbf{Distance Constraint:} $Dist(o_i) \le \tau_{dist}$ (e.g., 1.5 meters), preventing the inclusion of noisy or misaligned detections from distant observations.
\end{itemize}
Objects that pass these filters are registered in the \textit{Object Manager} with their 3D coordinates and associated zone IDs. In our approach, a zone is defined by the unique set of objects detected within a spatial cluster, rather than by architectural boundaries.

\subsection{Reasoning Layer: LLM Integration}
The reasoning layer serves as the ``cognitive engine'' that translates discrete object observations into high-level semantic contexts~\cite{yao2022react}.

\subsubsection{LoRA-based Fine-tuning}
To adapt a general-purpose LLM (Llama-2-7b-chat) to the specific spatial logic of indoor environments, we employ Low-Rank Adaptation (LoRA)~\cite{hu2021lora}. By fine-tuning on a dataset of object-zone co-occurrence patterns within AI2-THOR~\cite{kolve2017ai2thor}, the model learns to infer the most probable semantic zone category even from sparse visual cues. This domain-specific adaptation minimizes hallucinations and ensures that the inferred semantic priors are grounded in robotic navigation constraints, where a zone is characterized by the functional clusters of objects detected therein.

\subsubsection{Semantic Inference Process}
The Decision-Making Module (DMM) constructs a natural language prompt $P$ by verbalizing the set of filtered objects $O = \{o_1, o_2, \dots, o_n\}$ currently associated with the agent's location~\cite{wei2022cot}. The prompt is structured to elicit two key outputs:
\begin{enumerate}
\item \textbf{Zone Category ($Z_{est}$):} The semantic label of the current area defined by observed objects (e.g., ``Kitchen Area'').
\item \textbf{Target Existence Probability ($P_{target}$):} A scalar value $[0, 1]$ representing the likelihood of finding the target object in $Z_{est}$~\cite{ahn2022saycan}.
\end{enumerate}

These values are then integrated as attributes of the corresponding node in the topological map~\cite{thrun2005probabilistic}.

\subsection{Mapping Layer: Topological-Grid Hybrid Map}
A key contribution of this work is the dual-layer mapping approach that ensures both geometric precision and semantic abstraction~\cite{elfes1989occupancy}.

\subsubsection{Metric Layer (Occupancy Grid)}
The low-level metric map is an occupancy grid used for obstacle avoidance and local path planning~\cite{hart1968astar}. The agent updates this grid using depth sensor data at each step~\cite{deitke2020robothor}. We employ the A* algorithm over this grid to generate collision-free paths to specific coordinates within a zone or towards identified frontiers~\cite{chaplot2020topological}.

\subsubsection{Topological Layer (Semantic Graph)}
The high-level topological map represents the environment as a graph $\mathcal{G} = (\mathcal{V}, \mathcal{E})$~\cite{thrun2005probabilistic}. In our framework, this layer abstracts the environment into a collection of semantic zones rather than rigid architectural rooms.
\begin{itemize}
    \item \textbf{Nodes ($\mathcal{V}$):} Each node $v \in \mathcal{V}$ represents a distinct semantic zone. A new node is instantiated or updated whenever the agent detects a significant shift in the observed object set, which serves as a semantic signature for the current location~\cite{mccormac2017semanticfusion}.
    \item \textbf{Edges ($\mathcal{E}$):} Edges represent traversable connections between adjacent zones. This allows the agent to reason about zone-to-zone connectivity, effectively pruning the search space by bypassing zones with low $P_{target}$ based on their object-driven semantic descriptions~\cite{sunderhauf2017meaningful}.
\end{itemize}

\subsubsection{Object Manager}
The \textit{Object Manager} serves as the bridge between the two layers~\cite{rosinol2019kimera}. It stores each detected object $o_i$ as a tuple $(\mathbf{x}_i, l_i, v_{ID})$, where $\mathbf{x}_i$ is the 3D coordinate in the metric layer, $l_i$ is the semantic label, and $v_{ID}$ is the associated topological node ID (Zone ID)~\cite{stuckler2014semantic}. This structure enables the agent to remember ``what'' was found ``where,'' which is crucial for the verbalization process in subsequent reasoning steps~\cite{anderson2018vln}.

\section{EXPLORATION STRATEGY}
\label{sec:strategy}

The proposed exploration strategy integrates geometric completeness with semantic efficiency by prioritizing areas likely to contain the target object~\cite{yamauchi1997frontier}.

\subsection{Semantic Frontier Selection}
The agent identifies a set of frontiers $\mathcal{F} = \{f_1, f_2, \dots, f_m\}$, defined as the boundaries between known free space and unexplored regions in the occupancy grid~\cite{elfes1989occupancy}. Unlike standard frontier exploration, which selects the nearest $f_i$ based on Euclidean distance, our approach assigns a semantic weight $W(f_i)$ to each frontier:
\begin{equation}
W(f_i) = \alpha \cdot \frac{1}{D(a, f_i)} + \beta \cdot P_{target}(v \in \mathcal{V}_i)
\end{equation}~\cite{chaplot2020ans}
where $D(a, f_i)$ is the geodesic distance from the agent $a$ to the frontier, $P_{target}$ is the target existence probability inferred by the LLM for the semantic zone node $v$ containing $f_i$, and $\alpha, \beta$ are weighting coefficients~\cite{hart1968astar}. In our experiments, we set $\alpha = 1.0$ and $\beta = 0.5$ to ensure that the agent does not ignore nearby frontiers while still being guided by the LLM's semantic priors~\cite{hu2021lora}. This mechanism ensures that the agent prioritizes ``semantic-rich'' frontiers, such as unexplored corners of a kitchen-like zone when searching for a kettle, over irrelevant areas like a bathroom zone~\cite{kolve2017ai2thor}.

\subsection{Path Planning via Traveling Salesman Problem (TSP)}
Once a target zone node $v$ is selected, the agent must efficiently scan the area~\cite{thrun2005probabilistic}. To avoid redundant movement, we formulate the local scanning task as a Traveling Salesman Problem (TSP)~\cite{weihs2021pddl}. 
\begin{enumerate}
    \item \textbf{Candidate Generation:} A set of scanning points $\mathcal{P} = \{p_1, p_2, \dots, p_k\}$ is generated around the selected frontiers to ensure a $360^{\circ}$ field of view~\cite{savva2019habitat}.
    \item \textbf{Route Optimization:} The agent calculates the optimal visiting order by solving the TSP, minimizing the total path length $L = \sum L(p_j, p_{j+1})$, where $L(p_j, p_{j+1})$ is the distance computed by the A* algorithm on the grid map~\cite{deitke2020robothor}.
\end{enumerate}
This optimization allows the robot to systematically exhaust a high-probability zone before transitioning to the next location~\cite{chen2023how}.

\subsection{State Transition and Mode Switching}
The exploration process is governed by a finite state machine with three primary modes~\cite{pineau2003pbvi}:

\begin{itemize}
    \item \textbf{Local Exploration:} The agent executes TSP-optimized paths within the current semantic zone node $v_i$~\cite{silver2010montecarlo}.
    \item \textbf{Inter-zone Navigation:} When $\mathcal{F} \cap v_i = \emptyset$, or a neighbor node $v_j$ yields a significantly higher $P_{target}$, the agent navigates through a topological edge $e \in \mathcal{E}$ to the next node~\cite{bellman1957dynamic}.
    \item \textbf{Object Verification:} Triggered when a high-similarity object ($S \ge \tau_{target}$) is detected. The agent approaches the object to issue the final \texttt{Stop} command~\cite{jain2019learning}.
\end{itemize}
The detection of a significant change in the observed object set serves as a ``context reset'' signal, triggering a new LLM inference cycle to update the semantic priors for the newly identified zone~\cite{fried2018speaker}.

\section{EXPERIMENTS AND RESULTS}
\label{sec:experiments}
\figEnvironmentMaps

\subsection{Experimental Setup}
We evaluated the proposed framework in the AI2-THOR~\cite{kolve2017ai2thor} simulation environment, which provides photorealistic indoor scenes with rich object-zone correlations.

\begin{itemize}
\item \textbf{Environment:} 20 diverse scenes across four categories (Kitchen, Living Room, Bedroom, Bathroom) were used~\cite{savva2019habitat}. In our framework, these categories serve as initial semantic priors for the zones identified during exploration.
\item \textbf{Target Objects:} Objects typical of each zone (e.g., Kettle in a Kitchen-like zone, Remote in a Living Room-like zone) were designated as search targets~\cite{anderson2018evaluation}.
\item \textbf{Agent Configuration:} The agent was equipped with a monocular RGB-D camera and utilized ground-truth odometry. All visual processing and LLM inference were performed on an NVIDIA RTX 3090 GPU. In our implementation, we set the semantic weight parameters to $\alpha = 1.0$ and $\beta = 0.5$, and used object filtering thresholds of $\tau_{pixel} = 400$ and $\tau_{dist} = 1.5$m.
\end{itemize}

\subsection{Performance Metrics}
To assess search efficiency and reliability, we employed three standard metrics~\cite{deitke2020robothor}:
\begin{enumerate}
\item \textbf{Success Rate (SR):} The ratio of episodes where the agent successfully reaches the target within 1.0m~\cite{chaplot2020object}.
\item \textbf{Success weighted by Path Length (SPL):} Defined as $SPL = \frac{1}{N} \sum_{i=1}^{N} S_i \frac{L_i}{\max(P_i, L_i)}$, where $L_i$ is the shortest path and $P_i$ is the actual path.
\item \textbf{Total Distance (TD):} The average cumulative distance traveled per episode.
\end{enumerate}

\subsection{Baseline Comparisons}
The proposed method was compared against three baselines~\cite{anderson2018evaluation}:
\begin{itemize}
\item \textbf{Random Walk:} Moves randomly without mapping or reasoning~\cite{sutton2018reinforcement}.
\item \textbf{Standard Frontier (SF):} Exploration based solely on the nearest unvisited geometric boundary~\cite{yamauchi1997frontier}.
\item \textbf{Reactive LLM:} An LLM agent that decides the next action based on the current view without a global map or explicit zone-based memory~\cite{yao2022react}.
\end{itemize}

\subsection{Result Analysis and Discussion}
\figAgentObservationsLiving
\figExperimentalResults
Figure~\ref{fig:results} summarizes the performance across all scenarios~\cite{deitke2020robothor}. The proposed framework achieved the highest SR ($85\%$) and SPL ($0.52$), outperforming the SF baseline ($SPL=0.31$) and the Reactive LLM ($SR=40\%$).

\subsubsection{Ablation Study: Impact of LoRA Fine-tuning}
To analyze the contribution of domain-specific adaptation, we compared the fine-tuned Llama-2 (Proposed) with a zero-shot Llama-2-7b-chat~\cite{hu2021lora}.
\begin{itemize}
\item \textbf{Zone Inference Accuracy:} The LoRA-tuned model achieved an accuracy of $92\%$ in identifying semantic zone categories from object lists, whereas the zero-shot model frequently misidentified spaces due to unfamiliarity with AI2-THOR's specific object layouts and the functional definition of zones~\cite{kolve2017ai2thor}.
\item \textbf{Search Efficiency:} The zero-shot agent often engaged in redundant scanning in irrelevant zones, whereas the proposed agent successfully ``pruned'' low-probability zones based on their constituent objects, resulting in a $30\%$ reduction in total distance traveled.
\end{itemize}

\subsection{Discussion}
The results demonstrate that while geometric exploration (SF) ensures completeness, it lacks the efficiency provided by semantic priors~\cite{elfes1989occupancy}. Conversely, while LLMs provide priors, they require the structured memory of a topological map to avoid the myopic loops observed in the Reactive baseline~\cite{thrun2005probabilistic}. The synergy between LoRA-based semantic inference and hybrid mapping, where locations are represented as object-described zones, is thus critical for purposeful navigation~\cite{chen2023how}.


\section{CONCLUSION}
\label{sec:conclusion}

\subsection{Summary}
In this paper, we proposed a novel semantic navigation framework that integrates LLM-based commonsense reasoning with a hybrid topological-grid mapping system~\cite{yao2022react}. By transitioning from a ``reactive'' paradigm to a ``map-based'' AI approach, we addressed the fundamental limitations of traditional geometric exploration and myopic LLM-based agents~\cite{elfes1989occupancy}. 

Central to our approach is the concept of a ``zone''—a spatial unit defined and identified by the set of objects observed within it, rather than by physical boundaries. 

Our experimental results in the AI2-THOR environment demonstrated that fine-tuning a Large Language Model via LoRA significantly enhances the robot's ability to infer these semantic zone categories and target existence probabilities from sparse object observations~\cite{kolve2017ai2thor,hu2021lora}. The proposed hybrid map allows the agent to maintain long-term spatial consistency, enabling purposeful exploration that prioritizes semantic-rich areas defined by functional object clusters~\cite{thrun2005probabilistic}. Quantitative evaluations showed that our method achieves a superior Success Rate (SR) and Success weighted by Path Length (SPL) compared to traditional frontier exploration and reactive baselines~\cite{deitke2020robothor}.

\subsection{Future Work}
Future research will focus on the following three directions~\cite{sutton2018reinforcement}:
\begin{enumerate}
    \item \textbf{Dynamic Environment Adaptation:} We aim to extend the mapping layer to handle dynamic obstacles and human presence, allowing the topological graph of semantic zones to update its connectivity in real-time~\cite{rosinol2019kimera}.
    \item \textbf{Multi-Agent Collaboration:} Scaling the framework to multi-robot systems where semantic zone and topological information is shared to optimize search efficiency in large-scale environments~\cite{pineau2003pbvi}.
    \item \textbf{Multimodal Context Awareness:} Integrating visual and acoustic features beyond simple object labels into the LLM prompt to enable more nuanced situational reasoning and zone definition during exploration~\cite{liu2023llava}.
\end{enumerate}


\bibliographystyle{IEEEtran}
\bibliography{references}

\end{document}